\documentclass[letterpaper,twoside,english,journal,final,twocolumn]{IEEEtran}
\usepackage[T1]{fontenc}
\usepackage[latin1]{inputenc}
\usepackage{verbatim}
\usepackage{amsmath}
\usepackage{amssymb}
\usepackage{graphicx}

\makeatletter
\adddialect\l@ENGLISH\l@english

\usepackage{graphicx}

\makeatother

\usepackage{babel}
\begin{document}
\markboth{IEEE Transactions on Robotics}{IEEE Transactions on Robotics}

\title{Robust Recognition of Simultaneous Speech By a Mobile Robot}

\author{\authorblockN{Jean-Marc Valin\(^{*\diamond}\), Shun'ichi Yamamoto\(^\dag\),
Jean Rouat\(^\diamond\), \\
Fran\c{c}ois Michaud\(^\diamond\), Kazuhiro Nakadai\(^\ddag\), Hiroshi G.
Okuno\(^\dag\)}\\
\authorblockA{\textit{\(^*\)CSIRO ICT Centre, Sydney, Australia\\ \(^\diamond\)Department of Electrical Engineering and Computer Engineering \\ Universit\'{e} de Sherbrooke, 2500 boul. Universit\'{e}, Sherbrooke, Quebec, J1K 2R1, Canada\\ \(^\dag\)Graduate School of Informatics, Kyoto University, Sakyo, Kyoto, 606-8501, Japan\\ \(^\ddag\)HONDA Research Institute Japan Co., Ltd, 8-1 Honcho, Wako, Saitama, 351-0114, Japan\\ Jean-Marc.Valin@csiro.au\\ \{shunichi, okuno\}@kuis.kyoto-u.ac.jp\\ \{Jean.Rouat, Francois.Michaud\}@USherbrooke.ca\\nakadai@jp.honda-ri.com}}
\thanks{$^*$Corresponding author: Jean-Marc Valin (jean-marc.valin@csiro.au)\break
\copyright 2010 IEEE.  Personal use of this material is permitted. Permission from IEEE must be obtained for all other uses, in any current or future media, including reprinting/republishing this material for advertising or promotional purposes, creating new collective works, for resale or redistribution to servers or lists, or reuse of any copyrighted component of this work in other works.}
}
\maketitle
\begin{abstract}
This paper describes a system that gives a mobile robot the ability
to perform automatic speech recognition with simultaneous speakers.
A microphone array is used along with a real-time implementation of
Geometric Source Separation and a post-filter that gives a further
reduction of interference from other sources. The post-filter is also
used to estimate the reliability of spectral features and compute
a missing feature mask. The mask is used in a missing feature theory-based
speech recognition system to recognize the speech from simultaneous
Japanese speakers in the context of a humanoid robot. Recognition
rates are presented for three simultaneous speakers located at 2 meters
from the robot. The system was evaluated on a 200-word vocabulary
at different azimuths between sources, ranging from 10 to 90 degrees.
Compared to the use of the microphone array source separation alone,
we demonstrate an average reduction in relative recognition error
rate of 24\% with the post-filter and of 42\% when the missing features
approach is combined with the post-filter. We demonstrate the effectiveness
of our multi-source microphone array post-filter and the improvement
it provides when used in conjunction with the missing features theory.\end{abstract}

\section{Introduction\label{sec:Introduction}}

The human hearing sense is very good at focusing on a single source
of interest and following a conversation even when several people
are speaking at the same time. This ability is known as the \emph{cocktail
party effect} \cite{Cherry1953}. To operate in human and natural
settings, autonomous mobile robots should be able to do the same.
This means that a mobile robot should be able to separate and recognize
all sound sources present in the environment at any moment. This requires
the robots not only to detect sounds, but also to locate their origin,
separate the different sound sources (since sounds may occur simultaneously),
and process all of this data to be able to extract useful information
about the world from these sound sources. 

Recently, studies on robot audition have become increasingly active
\cite{Asano2001,ValinICRA2004,ValinIROS2004,Yamamoto04b,Yamamoto04a,Wang2004,Mungamuru2004}.
Most studies focus on sound source localization and separation. Recognition
of separated sounds has not been addressed as much, because it requires
integration of sound source separation capability with automatic speech
recognition, which is not trivial. Robust speech recognition usually
assumes source separation and/or noise removal from the feature vectors.
When several people speak at the same time, each separated speech
signal is severely distorted in spectrum from its original signal.
This kind of interference is more difficult to counter than background
noise because it is non-stationary and similar to the signal of interest.
Therefore, conventional noise reduction techniques such as spectral
subtraction \cite{Boll79}, used as a front-end of an automatic speech
recognizer, usually do not work well in practice. 

We propose the use of a microphone array and a sound source localization
system integrated with an automatic speech recognizer using the missing
feature theory \cite{Renevey01,Barker00} to improve robustness against
non-stationary noise. In previous work \cite{Yamamoto04b}, missing
feature theory was demonstrated using a mask computed from clean (non-mixed)
speech. The system we now propose can be used in a real environment
by computing the missing feature mask only from the data available
to the robot. To do so, a microphone array is used and a missing feature
mask is generated based only on the signals available from the array
post-filtering module. 

This paper focuses on the integration of speech/signal processing
and speech recognition techniques into a complete system operating
in a real (non-simulated) environment, demonstrating that such an
approach is functional and can operate in real-time. The novelty of
this approach lies in the way we estimate the missing feature mask
in the speech recognizer and in the tight integration of the different
modules. 

More specifically, we propose an original way of computing the missing
feature mask for the speech recognizer that relies on a measure of
frequency bins quality, estimated by our proposed postfilter. In opposition
to most missing feature techniques, our approach does not need estimation
of prior characteristics of the corrupting sources or noise. This
leads to new capabilities in robot speech recognition with simultaneous
speakers. As an example, for three simultaneous speakers, our system
can allow at least three speech recognizers running simultaneously
on the three separated speaker signals.

It is one of the first systems that runs in real-time on real robots
while performing simultaneous speech recognition. The real-time constraints
guided us in the integration of signal and speech processing techniques
that are sufficiently fast and efficient. We therefore had to reject
signal processing techniques that are too complex, even if potentially
yielding better performance.

The paper is organized as follows. Section \ref{sec:Speech-enhancement}
discusses the state of the art and limitations of speech enhancement
and missing feature-based speech recognition. Section \ref{sec:System-overview}
gives an overview of the system. Section \ref{sec:LSS} presents the
linear separation algorithm and Section \ref{sec:Loudness-domain-spectral-attenuation}
describes the proposed post-filter. Speech recognition integration
and computation of the missing feature mask are shown in Section \ref{sec:Integration-With-ASR}.
Results are presented in Section \ref{sec:Results}, followed by the
conclusion.

\section{Audition in Mobile Robotics\label{sec:Speech-enhancement}}

Artificial hearing for robots is a research topic still in its infancy,
at least when compared to the work already done on artificial vision
in robotics. However, the field of artificial audition has been the
subject of much research in recent years. In 2004, the IEEE/RSJ International
Conference on Intelligent Robots and Systems (IROS) included for the
first time a special session on robot audition. Initial work on sound
localization by Irie \cite{irie95robust} for the Cog \cite{Brooks99}
and Kismet robots can be found as early as 1995. The capabilities
implemented were however very limited, partly because of the necessity
to overcome hardware limitations.

The SIG robot\footnote{http://winnie.kuis.kyoto-u.ac.jp/SIG/oldsig/}
and its successor SIG2\footnote{http://winnie.kuis.kyoto-u.ac.jp/SIG/},
both developed at Kyoto University, have integrated increasing auditory
capabilities \cite{Nakadai00-AAAI,Nakadai,nakadai-hidai-okuno-kitano2001,OKUNO,nakadai-okuno-kitano2002,nakadai-okuno-kitano2002',nakadai-matsuura-okuno-kitano2003}
over the years (from 2000 to now). Both robots are based on binaural
audition, which is still the most common form of artificial audition
on mobile robots. Original work by Nakadai \emph{et al.} \cite{Nakadai00-AAAI,Nakadai}
on active audition has made it possible to locate sound sources in
the horizontal plane using binaural audition and active behavior to
disambiguate front from rear. Later work has focused more on sound
source separation \cite{nakadai-okuno-kitano2002,nakadai-okuno-kitano2002'}
and speech recognition \cite{Yamamoto04b,Yamamoto04a}. 

The ROBITA robot, designed at Waseda University, uses two microphones
to follow a conversation between two people, originally requiring
each participant to wear a headset \cite{matsusaka-tojo-kubota-furukawa-tamiya-hayata-nakano-kobayashi99},
although a more recent version uses binaural audition \cite{Matsusaka2001}.

A completely different approach is used by Zhang and Weng \cite{ZHANG}
in the SAIL robot with the goal of making a robot develop auditory
capabilities autonomously. In this case, the \emph{Q-learning} unsupervised
learning algorithm is used instead of supervised learning, which is
most commonly used in the field of speech recognition. The approach
is validated by making the robot learn simple voice commands. Although
current speech recognition accuracy using conventional methods is
usually higher than the results obtained, the advantage is that the
robot learns words autonomously.

More recently, robots have started taking advantage of using more
than two microphones. This is the case of the Sony QRIO SDR-4XII robot
\cite{Fujita03} that features seven microphones. Unfortunately, little
information is available regarding the processing done with those
microphones. A service robot by Choi \emph{et al.} \cite{choi-kong-kim-bang2003}
uses eight microphones organized in a circular array to perform speech
enhancement and recognition. The enhancement is provided by an adaptive
beamforming algorithm. Work by Asano, Asoh, \emph{et al.} \cite{Asano2001,Asoh97,Asano99}
also uses a circular array composed of eight microphones on a mobile
robot to perform both localization and separation of sound sources.
In more recent work \cite{Asoh2004}, particle filtering is used to
integrate vision and audition in order to track sound sources.

In general, human-robot interface is a popular area of audition-related
research in robotics. Works on robot audition for human-robot interface
has also been done by Prodanov \emph{et al.} \cite{Prodanov} and
Theobalt \emph{et al.} \cite{Theobalt}, based on a single microphone
near the speaker. Even though human-robot interface is the most common
goal of robot audition research, there is research being conducted
for other goals. Huang \emph{et al.} \cite{Huang1999Navig} use binaural
audition to help robots navigate in their environment, allowing a
mobile robot to move toward sound-emitting objects without colliding
with those object. The approach even works when those objects are
not visible (i.e., not in line of sight), which is an advantage over
vision.

\section{System overview\label{sec:System-overview}}

One goal of the proposed system is to integrate the different steps
of source separation, speech enhancement and speech recognition as
closely as possible to maximize recognition accuracy by using as much
of the available information as possible and with a strong real-time
constraint. We use a microphone array composed of omni-directional
elements mounted on the robot. The missing feature mask is generated
in the time-frequency plane since the separation module and the post-filter
already use this signal representation. We assume that all sources
are detected and localized by an algorithm such as \cite{ValinICASSP2006,Valin2007},
although our approach is not specific to any localization algorithm.
The estimated location of the sources is used by a linear separation
algorithm. The separation algorithm we use is a modified version of
the Geometric Source Separation (GSS) approach proposed by Parra and
Alvino \cite{Parra2002Geometric}, designed to suit our needs for
real-time and real-life applications. We show that it is possible
to implement the separation with relatively low complexity that grows
linearly with the number of microphones. The method is interesting
for use in the mobile robotics context because it makes it easy to
dynamically add or remove sound sources as they appear or disappear.
The output of the GSS still contains residual background noise and
interference, that we further attenuate through a multi-channel post-filter.
The novel aspect of this post-filter is that, for each source of interest,
the noise estimate is decomposed into stationary and transient components
assumed to be due to leakage between the output channels of the initial
separation stage. In the results, the performance of that post-filter
is shown to be superior to those obtained when considering each separated
source independently.

The post-filter we use can not only reduce the amount of noise and
interference, but its behavior provides useful information that is
used to evaluate the reliability of different regions of the time-frequency
plane for the separated signals. Based also on the ability of the
post-filter to model independently background noise and interference,
we propose a novel way to estimate the missing feature mask to further
improve speech recognition accuracy. This also has the advantage that
acoustic models trained on clean data can be used and that no multi-condition
training is required.

\begin{figure}
\includegraphics[width=1\columnwidth]{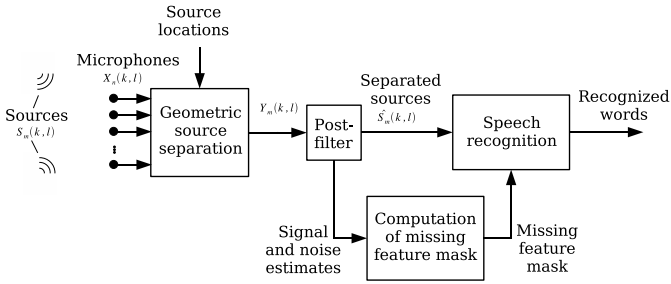}

\caption{Overview of the separation system with the post-filter being used
both to improve the audio quality and to estimate the missing feature
mask.\label{cap:Overview}}
\end{figure}

The structure of the proposed system is shown in Fig. \ref{cap:Overview}
and its four main parts are:
\begin{enumerate}
\item Linear separation of the sources, implemented as a variant of the
Geometric Source Separation (GSS) algorithm; 
\item Multi-channel post-filtering of the separated output; 
\item Computation of the missing feature mask from the post-filter output; 
\item Speech recognition using the separated audio and the missing feature
mask. 
\end{enumerate}

\section{Geometric Source Separation\label{sec:LSS}}

Although the work we present can be adapted to systems with any linear
source separation algorithm, we propose to use the Geometric Source
Separation (GSS) algorithm because it is simple and well suited to
a mobile robotics application. More specifically, the approach has
the advantage that it can make use of the location of the sources.
In this work, we only make use of the direction information, which
can be obtained with a high degree of accuracy using the method described
in \cite{ValinICRA2004}. It was shown in \cite{ValinICASSP2006}
that distance can be estimated as well. The use of location information
is important when new sources are observed. In that situation, the
system can still provide acceptable separation performance (at least
equivalent to the delay-and-sum beamformer), even if the adaptation
has not yet taken place.

The method operates in the frequency domain using a frame length of
21 ms (1024 samples at 48 kHz). Let $S_{m}(k,\ell)$ be the real (unknown)
sound source $m$ at time frame $\ell$ and for discrete frequency
$k$. We denote as $\mathbf{s}(k,\ell)$ the vector of the sources
$S_{m}(k,\ell)$ and matrix $\mathbf{A}(k)$ as the transfer function
from the sources to the microphones. The signal received at the microphones
is thus given by:
\begin{equation}
\mathbf{x}(k,\ell)=\mathbf{A}(k)\mathbf{s}(k,\ell)+\mathbf{n}(k,\ell)\label{eq:GSS_mixing}
\end{equation}
 where $\mathbf{n}(k,\ell)$ is the non-coherent background noise
received at the microphones. The matrix $\mathbf{A}(k)$ can be estimated
using the result of a sound localization algorithm by assuming that
all transfer functions have unity gain and that no diffraction occurs.
The elements of $\mathbf{A}(k)$ are thus expressed as:
\begin{equation}
a_{ij}(k)=e^{-\jmath2\pi k\delta_{ij}}\label{eq:GSS_a_ij}
\end{equation}
 where $\delta_{ij}$ is the time delay (in samples) to reach microphone
$i$ from source $j$.

The separation result is then defined as $\mathbf{y}(k,\ell)=\mathbf{W}(k,\ell)\mathbf{x}(k,\ell)$,
where $\mathbf{W}(k,\ell)$ is the separation matrix that must be
estimated. This is done by providing two constraints (the index $\ell$
is omitted for the sake of clarity):
\begin{enumerate}
\item Decorrelation of the separation algorithm outputs (second order statistics
are sufficient for non-stationary sources), expressed as $\mathbf{R}_{\mathbf{yy}}(k)-\mathrm{diag}\left[\mathbf{R}_{\mathbf{yy}}(k)\right]=\mathbf{0}$. 
\item The geometric constraint $\mathbf{W}(k)\mathbf{A}(k)=\mathbf{I}$,
which ensures unity gain in the direction of the source of interest
and places zeros in the direction of interferences. 
\end{enumerate}
In theory, constraint 2) could be used alone for separation (the method
is referred to as LS-C2 \cite{Parra2002Geometric}), but this is insufficient
in practice, as the method does not take into account reverberation
or errors in localization. It is also subject to instability if $\mathbf{A}(k)$
is not invertible at a specific frequency. When used together, constraints
1) and 2) are too strong. For this reason, we use a ``soft'' constraint
(refereed to as GSS-C2 in \cite{Parra2002Geometric}) combining 1)
and 2) in the context of a gradient descent algorithm.

Two cost functions are created by computing the square of the error
associated with constraints 1) and 2). These cost functions are defined
as, respectively:
\begin{align}
J_{1}(\mathbf{W}(k)) & =\left\Vert \mathbf{R}_{\mathbf{yy}}(k)-\mathrm{diag}\left[\mathbf{R}_{\mathbf{yy}}(k)\right]\right\Vert ^{2}\label{eq:Cost_J1}\\
J_{2}(\mathbf{W}(k)) & =\left\Vert \mathbf{W}(k)\mathbf{A}(k)-\mathbf{I}\right\Vert ^{2}\label{eq:Cost_J2}
\end{align}
 where the matrix norm is defined as $\left\Vert \mathbf{M}\right\Vert ^{2}=\mathrm{trace}\left[\mathbf{M}\mathbf{M}^{H}\right]$
and is equal to the sum of the square of all elements in the matrix.
The gradient of the cost functions with respect to $\mathbf{W}(k)$
is equal to \cite{Parra2002Geometric}:
\begin{align}
\frac{\partial J_{1}(\mathbf{W}(k))}{\partial\mathbf{W}^{*}(k)} & =4\mathbf{E}(k)\mathbf{W}(k)\mathbf{R}_{\mathbf{xx}}(k)\label{eq:Grad_J1}\\
\frac{\partial J_{2}(\mathbf{W}(k))}{\partial\mathbf{W}^{*}(k)} & =2\left[\mathbf{W}(k)\mathbf{A}(k)-\mathbf{I}\right]\mathbf{A}(k)\label{eq:Grad_J2}
\end{align}
where $\mathbf{E}(k)=\mathbf{R}_{\mathbf{yy}}(k)-\mathrm{diag}\left[\mathbf{R}_{\mathbf{yy}}(k)\right]$.

The separation matrix $\mathbf{W}(k)$ is then updated as follows:
\begin{equation}
\mathbf{W}^{n+1}(k)=\mathbf{W}^{n}(k)-\mu\!\left[\alpha(k)\frac{\partial J_{1}(\mathbf{W}(k))}{\partial\mathbf{W}^{*}(k)}\!+\!\frac{\partial J_{2}(\mathbf{W}(k))}{\partial\mathbf{W}^{*}(k)}\right]\label{eq:W_update}
\end{equation}
where $\alpha(f)$ is an energy normalization factor equal to $\left\Vert \mathbf{R}_{\mathbf{xx}}(k)\right\Vert ^{-2}$
and $\mu$ is the adaptation rate.

The difference between our implementation and the original GSS algorithm
described in \cite{Parra2002Geometric} lies in the way the correlation
matrices $\mathbf{R}_{\mathbf{xx}}(k)$ and $\mathbf{R}_{\mathbf{yy}}(k)$
are computed. Instead of using several seconds of data, our approach
uses instantaneous estimates, as used in the stochastic gradient adaptation
of the Least Mean Square (LMS) adaptive filter \cite{HaykinAFT}.
We thus assume that:
\begin{align}
\mathbf{R}_{\mathbf{xx}}(k) & =\mathbf{x}(k)\mathbf{x}(k)^{H}\label{eq:Rxx_instant}\\
\mathbf{R}_{\mathbf{yy}}(k) & =\mathbf{y}(k)\mathbf{y}(k)^{H}\label{eq:Ryy_instant}
\end{align}

It is then possible to rewrite (\ref{eq:Grad_J1}) as:
\begin{equation}
\frac{\partial J_{1}(\mathbf{W}(k))}{\partial\mathbf{W}^{*}(k)}=4\left[\mathbf{E}(k)\mathbf{W}(k)\mathbf{x}(k)\right]\mathbf{x}(k)^{H}\label{eq:Grad_J1_instant}
\end{equation}
which only requires matrix-by-vector products, greatly reducing the
complexity of the algorithm. Similarly, the normalization factor $\alpha(k)$
can also be simplified as $\left[\left\Vert \mathbf{x}(k)\right\Vert ^{2}\right]^{-2}$.
With a small update rate, it means that the time averaging is performed
implicitly. In early experiments, the instantaneous estimate of the
correlation was found to have no significant impact on the performance
of the separation, but is necessary for real-time implementation.

The weight initialization we use corresponds to a delay-and-sum beamformer,
referred to as the I1 (or C1) initialization method in \cite{Parra2002Geometric}.
Such initialization ensures that prior to adaptation, the performances
are at worst equivalent to a delay-and-sum beamformer. In fact, if
only a single source is present, our algorithm is strictly equivalent
to a delay-and-sum beamformer implemented in the frequency domain.

\section{Multi-channel post-filter\label{sec:Loudness-domain-spectral-attenuation}}

\begin{figure}
\includegraphics[width=1\columnwidth,keepaspectratio]{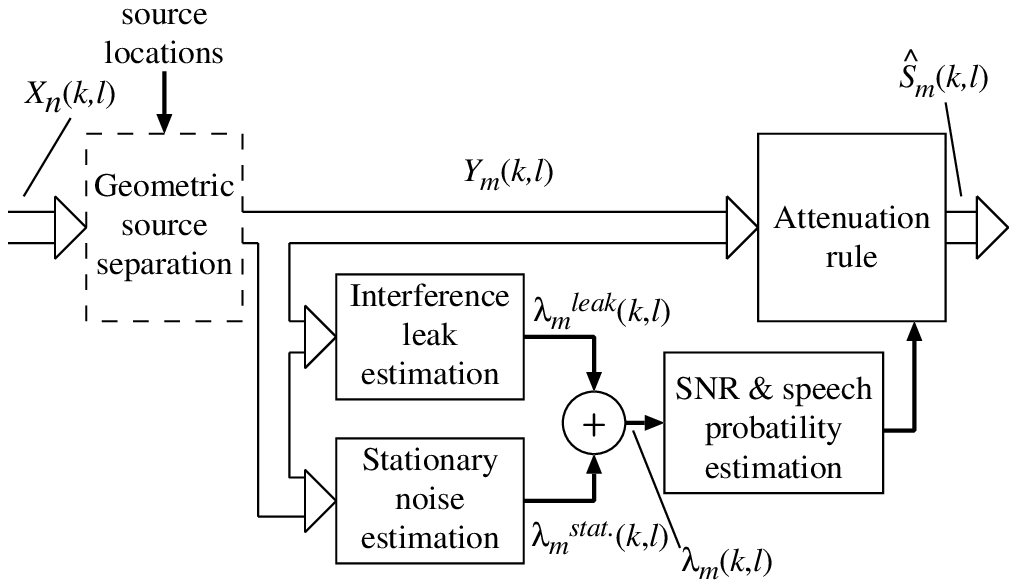}

\caption{Overview of the post-filter.\protect \\
 $X_{n}(k,\ell),n=0\ldots N-1$: Microphone inputs, $Y_{m}(k,\ell),\:m=0\ldots M-1$:
Inputs to the post-filter, $\hat{S}_{m}(k,\ell)=G_{m}(k,\ell)Y_{m}(k,\ell),\:m=0\ldots M-1$:
Post-filter outputs.\label{cap:Overview-Post-filter}}
\end{figure}

To enhance the output of the GSS algorithm presented in Section \ref{sec:LSS},
we derive a frequency-domain post-filter that is based on the optimal
estimator originally proposed by Ephraim and Malah \cite{EphraimMalah1984,EphraimMalah1985}.
Several approaches to microphone array post-filtering have been proposed
in the past. Most of these post-filters address reduction of stationary
background noise \cite{zelinski-1988,mccowan-icassp2002}. Recently,
a multi-channel post-filter taking into account non-stationary interferences
was proposed by Cohen \cite{CohenArray2002}. The novelty of our post-filter
resides in the fact that, for a given channel output of the GSS, the
transient components of the corrupting sources are assumed to be due
to leakage from the other channels during the GSS process. Furthermore,
for a given channel, the stationary and the transient components are
combined into a single noise estimator used for noise suppression,
as shown in Fig. \ref{cap:Overview-Post-filter}. In addition, we
explore different suppression criteria ($\alpha$ values) for optimizing
speech recognition instead of perceptual quality. Again, when only
one source is present, this post-filter is strictly equivalent to
standard single-channel noise suppression techniques.

\subsection{Noise Estimation\label{sub:Noise-estimation}}

This section describes the estimation of noise variances that are
used to compute the weighting function $G_{m}(k,\ell)$ by which the
outputs $Y_{m}(k,\ell)$ of the GSS is multiplied to generate a cleaned
signal whose spectrum is denoted $\hat{S}_{m}(k,\ell)$. The noise
variance estimation $\lambda_{m}(k,\ell)$ is expressed as:
\begin{equation}
\lambda_{m}(k,\ell)=\lambda_{m}^{stat.}(k,\ell)+\lambda_{m}^{leak}(k,\ell)\label{eq:noise_estim}
\end{equation}
 where $\lambda_{m}^{stat.}(k,\ell)$ is the estimate of the stationary
component of the noise for source $m$ at frame $\ell$ for frequency
$k$, and $\lambda_{m}^{leak}(k,\ell)$ is the estimate of source
leakage.

We compute the stationary noise estimate $\lambda_{m}^{stat.}(k,\ell)$
using the Minima Controlled Recursive Average (MCRA) technique proposed
by Cohen \cite{CohenNonStat2001}.

To estimate $\lambda_{m}^{leak}$ we assume that the interference
from other sources has been reduced by a factor $\eta$ (typically
$-10\:\mathrm{dB}\leq\eta\leq-3\:\mathrm{dB}$) by the separation
algorithm (GSS). The leakage estimate is thus expressed as:
\begin{equation}
\lambda_{m}^{leak}(k,\ell)=\eta\sum_{i=0,i\neq m}^{M-1}Z_{i}(k,\ell)\label{eq:leak_estim}
\end{equation}
 where $Z_{m}(k,\ell)$ is the smoothed spectrum of the $m^{th}$
source, $Y_{m}(k,\ell)$, and is recursively defined (with $\alpha_{s}=0.7$)
as:
\begin{equation}
Z_{m}(k,\ell)=\alpha_{s}Z_{m}(k,\ell-1)+(1-\alpha_{s})\left|Y_{m}(k,\ell)\right|^{2}\label{eq:smoothed_spectrum}
\end{equation}
It is worth noting that if $\eta=0$ or $M=1$, then the noise estimate
becomes $\lambda_{m}(k,\ell)=\lambda_{m}^{stat.}(k,\ell)$ and our
multi-source post-filter is reduced to a single-source post-filter.

\subsection{Suppression Rule}

From here on, unless otherwise stated, the $m$ index and the $\ell$
arguments are omitted for clarity and the equations are given for
each $m$ and for each $\ell$. The proposed noise suppression rule
is based on minimum mean-square error (MMSE) estimation of the spectral
amplitude in the ($\left|X(k)\right|^{\alpha}$) domain. The power
coefficient $\alpha$ is chosen to maximize the recognition results.

Assuming that speech is present, the spectral amplitude estimator
is defined by:
\begin{equation}
\hat{A}(k)=\left(E\left[\left|S(k)\right|^{\alpha}\left|Y(k)\right.\right]\right)^{\frac{1}{\alpha}}=G_{H_{1}}(k)\left|Y(k)\right|\label{eq:amplitude_estim}
\end{equation}
 where $G_{H_{1}}(k)$ is the spectral gain assuming that speech is
present.

The spectral gain for arbitrary $\alpha$ is derived from Equation
13 in \cite{EphraimMalah1985}:
\begin{equation}
G_{H_{1}}(k)=\frac{\sqrt{\upsilon(k)}}{\gamma(k)}\left[\Gamma\left(1+\frac{\alpha}{2}\right)M\left(-\frac{\alpha}{2};1;-\upsilon(k)\right)\right]^{\frac{1}{\alpha}}\label{eq:Gain_H1}
\end{equation}
 where $M(a;c;x)$ is the confluent hypergeometric function, $\gamma(k)\triangleq\left|Y(k)\right|^{2}/\lambda(k)$
and $\xi(k)\triangleq E\left[\left|S(k)\right|^{2}\right]/\lambda(k)$
are respectively the \emph{a posteriori} SNR and the \emph{a priori}
SNR. We also have $\upsilon(k)\triangleq\gamma(k)\xi(k)/\left(\xi(k)+1\right)$
\cite{EphraimMalah1984}.

The \emph{a priori} SNR $\xi(k)$ is estimated recursively as \cite{EphraimMalah1984}:
\begin{align}
\hat{\xi}(k,\ell) & =\alpha_{p}G_{H_{1}}^{2}(k,\ell-1)\gamma(k,\ell-1)\nonumber \\
 & +(1-\alpha_{p})\max\left\{ \gamma(k,\ell)-1,0\right\} \label{eq:xi_decision_directed}
\end{align}

When taking into account the probability of speech presence, we obtain
the modified spectral gain:
\begin{equation}
G(k)=p^{1/\alpha}(k)G_{H_{1}}(k)\label{eq:optimal_gain_final}
\end{equation}
where $p(k)$ is the probability that speech is present in the frequency
band $k$ and given by:
\begin{equation}
p(k)=\left\{ 1+\frac{\hat{q}(k)}{1-\hat{q}(k)}\left(1+\xi(k)\right)\exp\left(-\upsilon(k)\right)\right\} ^{-1}\label{eq:def_speech_prob}
\end{equation}

The \emph{a priori} probability of speech presence $\hat{q}(k)$ is
computed as in \cite{CohenNonStat2001} using speech measurements
on the current frame for a local frequency window, a larger frequency
and for the whole frame.

\section{Integration With Speech Recognition\label{sec:Integration-With-ASR}}

Robustness against noise in conventional\footnote{We use conventional in the sense of speech recognition for applications
where a single microphone is used in a static environment such as
a vehicle or an office.} automatic speech recognition (ASR) is being extensively studied,
in particular, in the AURORA project \cite{AURORA,Pearce01}. To realize
noise-robust speech recognition, \emph{multi-condition training} (training
on a mixture of clean speech and noises) has been studied \cite{Lippmann87,Blanchet92}.
This is currently the most common method for vehicle and telephone
applications. Because an acoustic model obtained by multi-condition
training reflects all expected noises in specific conditions, recognizer's
use of the acoustic model is effective as long as the noise is stationary.
This assumption holds for example with background noise in a vehicle
and on a telephone. However, multi-condition training is not effective
for mobile robots, since those usually work in dynamically changing
noisy environments and furthermore multi-condition training requires
an important amount of data to learn from. 

Source separation and speech enhancement algorithms for robust recognition
are another potential alternative for automatic speech recognition
on mobile robots. However, their common use is to maximize the perceptual
quality of the resulting signal. This is not always effective since
most preprocessing source separation and speech enhancement techniques
distort the spectrum and consequently degrade features, reducing the
recognition rate (even if the signal is perceived to be cleaner by
na\"{\i}ve listeners~\cite{OShaughnessy2003}). For example, the
work of Seltzer \emph{et al.} \cite{seltzer2003} on microphone arrays
addresses the problem of optimizing the array processing specifically
for speech recognition (and not for a better perception). Recently,
Araki \emph{et al.} \cite{Araki2004} have applied ICA to the separation
of three sources using only two microphones. Aarabi and Shi \cite{AARABI2004}
have shown speech enhancement feasibility, for speech recognition,
using only the phase of the signals from an array of microphones.

\subsection{Missing Features Theory and Speech Recognition}

Research of confident islands in the time-frequency plane representation
has been shown to be effective in various applications and can be
implemented with different strategies. One of the most effective is
the missing feature strategy. Cooke \textit{et al.}~\cite{cooke1994,cooke2001b}
propose a probabilistic estimation of a mask in regions of the time-frequency
plane where the information is not reliable. Then, after masking,
the parameters for speech recognition are generated and can be used
in conventional speech recognition systems. They obtain a significant
increase in recognition rates without any explicit modeling of the
noise~\cite{barker2001}. In this scheme, the mask is essentially
based on the dominance speech/interference criteria and a probabilistic
estimation of the mask is used.

Conventional missing feature theory based ASR is a Hidden Markov Model
(HMM) based recognizer, which output probability (emission probability)
is modified to keep only the reliable feature distributions. According
to the work by Cooke \textit{et al.}~\cite{cooke2001b}, HMMs are
trained on clean data. Density in each state $S_{i}$ is modeled using
mixtures of $M$ Gaussians with diagonal-only covariance.

Let $f(\mathbf{x}|S)$ be the output probability density of feature
vector $\mathbf{x}$ in state $S_{i}$, and $P(j|S_{i})$ represent
the mixture coefficients expressed as a probability. The output probability
density is defined by: 
\begin{equation}
f(\mathbf{x}|S_{i})=\sum_{j=1}^{M}P(j|S_{i})f(\mathbf{x}|j,S_{i})\label{eq:missingFeatre}
\end{equation}

Cooke \textit{et al.}~\cite{cooke2001b} propose to transform (\ref{eq:missingFeatre})
to take into consideration the only reliable features $x_{r}$ from
$\mathbf{x}$ and to remove the unreliable features. This is equivalent
to using the marginalization probability density functions $f(x_{r}|j,S_{i})$
instead of $f(\mathbf{x}|j,S_{i})$ by simply implementing a binary
mask. Consequently, only reliable features are used in the probability
calculation, and the recognizer can avoid undesirable effects due
to unreliable features.

Hugo van Hamme~\cite{vanhamme2003} formulates the missing feature
approach for speech recognizers using conventional parameters such
as mel frequency cepstral coefficients (MFCC). He uses data imputation
according to Cooke~\cite{cooke2001b} and proposes a suitable transformation
to be used with MFCC for missing features. The acoustic model evaluation
of the unreliable features is modified to express that their clean
values are unknown or confined within bounds. In a more recent paper,
Hugo van Hamme~\cite{vanhamme2004} presents speech recognition results
by integrating harmonicity in the signal to noise ratio for noise
estimation. He uses only static MFCC as, according to his observations,
dynamic MFCC do not increase sufficiently the speech recognition rate
when used in the context of missing features framework. The need to
estimate pitch and voiced regions in the time-space representation
is a limit to this approach. In a similar approach, Raj, Seltzer and
Stern~\cite{raj2004} propose to modify the spectral representation
to derive cepstral vectors. They present two missing feature algorithms
that reconstruct spectrograms from incomplete noisy spectral representations
(\textit{masked} representations). Cepstral vectors can be derived
from the reconstructed spectrograms for missing feature recognition.
Seltzer \textit{et al.}~\cite{seltzer2004SpeechComm} propose the
use of a Bayesian classifier to determine the reliability of spectrographic
elements. Ming, Jancovic and Smith~\cite{ming2002,ming2003} propose
the \textit{probabilistic union model} as an alternative to the missing
feature framework. According to the authors, methods based on the
missing feature framework usually require the identification of the
noisy bands. This identification can be difficult for noise with unknown,
time-varying band characteristics. They designed an approach for speech
recognition involving partial, unknown corrupted frequency-bands.
In their approach, they combine the local frequency-band information
based on the union of random events, to reduce the dependence of the
model on information about the noise. Cho and Oh~\cite{cho2004}
apply the \textit{union model} to improve robust speech recognition
based on frequency bands selection. From this selection, they generate
``channel-attentive'' mel frequency cepstral coefficients. Even
if the use of missing features for robust recognition is relatively
recent, many applications have already been designed.

To avoid the use of multi-condition training, we propose to merge
a multi-microphone source separation and speech enhancement system
with the missing feature approach. Very little work has been done
with arrays of microphones in the context of missing feature theory.
To our knowledge, only McCowan \emph{et al.} \cite{mccowan-rr-02-09-proc}
apply the missing feature framework to microphone arrays. Their approach
defines a missing feature mask based on the input-to-output ratio
of a post-filter but is however only validated on stationary noise. 

Some missing feature mask techniques can also require the estimation
of prior characteristics of the corrupting sources or noise. They
usually assume that the noise or interference characteristics vary
slowly with time. This is not possible in the context of a mobile
robot. We propose to estimate quasi-instantaneously the mask (without
preliminary training) by exploiting the post-filter outputs along
with the local gains (in the time-frequency plane representation)
of the post-filter. These local gains are used to generate the missing
feature mask. Thus, the speech recognizer with clean acoustic models
can adapt to the distorted sounds by consulting the post-filter feature
missing masks. This approach is also a solution to the automatic generation
of simultaneous missing feature masks (one for each speaker). It allows
the use of simultaneous speech recognizers (one for each separated
sound source) with their own mask.

\subsection{Reliability estimation}

The post-filter uses adaptive spectral estimation of background noise
and interfering sources to enhance the signal produced during the
initial separation. The main idea lies in the fact that, for each
source of interest, the noise estimate is decomposed into stationary
and transient components assumed to be due to leakage between the
output channels of the initial separation stage. It also provides
useful information concerning the amount of noise present at a certain
time, for each particular frequency. Hence, we use the post-filter
to estimate a missing feature mask that indicates how reliable each
spectral feature is when performing recognition.

\subsection{Computation of Missing Feature Masks\label{sec:Computation-of-Mask}}

The missing feature mask is a matrix representing the reliability
of each feature in the time-frequency plane. More specifically, this
reliability is computed for each frame and for each mel-frequency
band. This reliability can be either a continuous value from 0 to
1, or a discrete value of 0 or 1. In this paper, discrete masks are
used. It is worth mentioning that computing the mask in the mel-frequency
band domain means that it is not possible to use MFCC features, since
the effect of the DCT cannot be applied to the missing feature mask. 

For each mel-frequency band, the feature is considered reliable if
the ratio of the post-filter output energy over the input energy is
greater than a threshold $T$. The reason for this choice is that
it is assumed that the more noise is present in a certain frequency
band, the lower the post-filter gain will be for that band. 

One of the dangers of computing missing feature masks based on a signal-to-noise
measure is that there is a tendency to consider all silent periods
as non-reliable, because they are dominated by noise. This leads to
large time-frequency areas where no information is available to the
ASR, preventing it from correctly identifying silence (we made this
observation from practice). For this reason, it is desirable to consider
as reliable at least some of the silence, especially when there is
no non-stationary interference.

The missing feature mask is computed in two steps: for each frame
$\ell$ and for each mel frequency band $i$: 
\begin{enumerate}
\item We compute a continuous mask $m_{\ell}(i)$ that reflects the reliability
of the band: 
\begin{equation}
m_{\ell}(i)=\frac{S_{\ell}^{out}(i)+N_{\ell}(i)}{S_{\ell}^{in}(i)}\label{eq:cont_mask}
\end{equation}

where $S_{\ell}^{in}(i)$ and $S_{\ell}^{out}(i)$ are respectively
the post-filter input and output energy for frame $\ell$ at mel-frequency
band $i$, and $N_{\ell}(i)$ is the background noise estimate. The
values $S_{\ell}^{in}(i)$, $S_{\ell}^{out}(i)$ and $N_{\ell}(i)$
are computed using a mel-scale filterbank with triangular bandpass
filters, based on linear-frequency post-filter data.

\item We deduce a binary mask $M_{\ell}(i)$. This mask will be used to
remove the unreliable mel frequency bands at frame $\ell$:

\begin{equation}
M_{\ell}(i)=\left\{ \begin{array}{ll}
1, & m_{\ell}(i)>T\\
0,\quad & \mathrm{otherwise}
\end{array}\right.\label{eq:mask_threshold}
\end{equation}
 where $T$ is the mask threshold. We use the value $T=0.25$, which
produces the best results over a range of experiments. In practice
the algorithm is not very sensitive to $T$ and all values in the
$\left[0.15,0.30\right]$ interval generally produce equivalent results.

\end{enumerate}
In comparison to McCowan \emph{et al.} \cite{mccowan-rr-02-09-proc},
the use of the multi-source post-filter allows a better reliability
estimation by distinguishing between interference and background noise.
We include the background noise estimate $N_{\ell}(i)$ in the numerator
of (\ref{eq:cont_mask}) to ensure that the missing feature mask equals
1 when no speech source is present (as long as there is no interference).
Using a more conventional post-filter as proposed by McCowan \emph{et
al.} \cite{mccowan-rr-02-09-proc} and Cohen \emph{et al.} \cite{CohenArray2002}
would not allow the mask to preserve silence features, which is known
to degrade ASR accuracy. The distinction between background noise
and interference also reflects the fact that background noise cancellation
is generally more efficient than interference cancellation. 

\begin{figure*}
\begin{center}\includegraphics[width=0.75\paperwidth,keepaspectratio]{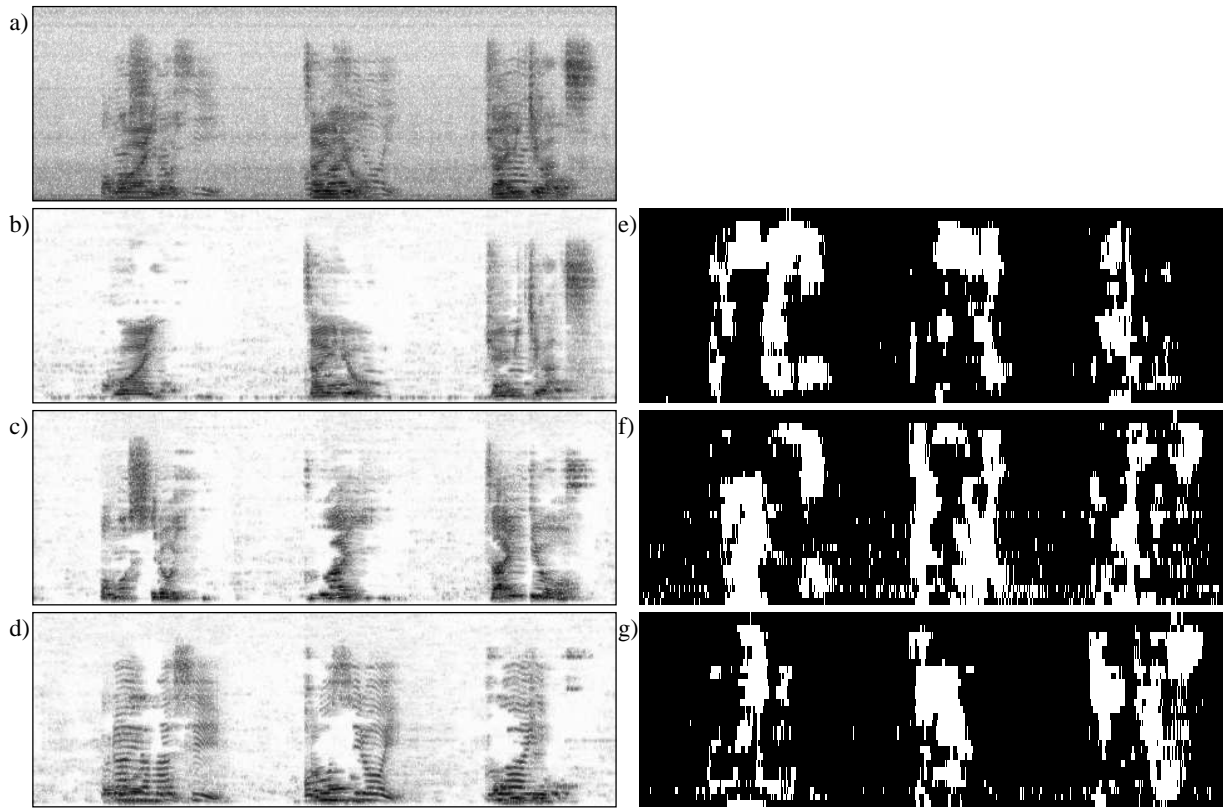}\end{center}

\caption{Spectrograms for separation of three speakers, $90^{\circ}$ apart
with post-filter. a) signal as captured at microphone \#1 b) separated
right speaker c) separated center speaker d) separated left speaker.
e)~-~g) corresponding mel-frequency missing feature mask for static
features with \textbf{reliable features ($M_{\ell}(i)=1$) shown in
black}. Time is represented on the \emph{x-axis} and frequency (0-8
kHz) on the \emph{y-axis}.\label{cap:Spectrogram-for-separation}}
\end{figure*}

An example of a computed missing feature mask is shown in Fig.~\ref{cap:Spectrogram-for-separation}.
It is observed that the mask indeed preserves the silent periods and
considers unreliable the regions of the spectrum dominated by other
sources. The missing feature mask for delta-features is computed using
the mask for the static features. The dynamic mask $\Delta M_{\ell}(i)$
is computed as:
\begin{equation}
\Delta M_{\ell}(i)=\prod_{k=-2}^{2}M_{\ell-k}(i)\label{eq:delta_mask}
\end{equation}
 and is non-zero only when all the mel features used to compute the
delta-cepstrum are deemed reliable.

\subsection{Speech Analysis for Missing Feature Masks }

Since MFCC cannot be easily used directly with a missing feature mask
and as the post-filter gains are expressed in the time--frequency
plane, we use spectral features that are derived from MFCC features
with the Inverse Discrete Cosine Transform (IDCT). The detailed steps
for feature generation are as follows:
\begin{enumerate}
\item {[}FFT{]} The speech signal sampled at 16 kHz is analyzed using a
400-sample FFT with a 160-sample frame shift. 
\item {[}Mel{]} The spectrum is analyzed by a $24^{\mathrm{th}}$ order
mel-scale filter bank. 
\item {[}Log{]} The mel-scale spectrum of the $24^{\mathrm{th}}$ order
is converted to log-energies. 
\item {[}DCT{]} The log mel-scale spectrum is converted by Discrete Cosine
Transform to the Cepstrum. 
\item {[}Lifter{]} Cepstral features 0 and 13-23 are set to zero so as to
make the spectrum smoother. 
\item {[}CMS{]} Convolutive effects are removed using Cepstral Mean Subtraction. 
\item {[}IDCT{]} The normalized Cepstrum is transformed back to the log
mel-scale spectral through an Inverse DCT. 
\item {[}Differentiation{]} The features are differentiated in the time,
producing 24 delta features in addition to the static features. 
\end{enumerate}
The {[}CMS{]} step is necessary to remove the effect of convolutive
noise, such as reverberation and microphone frequency response.

The same features are used for training and evaluation. Training is
performed on clean speech, without any effect from the post-filter.
In practice, this means that the acoustic model does not need to be
adapted in any way to our method. During evaluation, the only difference
with a conventional ASR is the use of the missing feature mask as
represented in (\ref{eq:missingFeatre}).

\subsection{The Missing Feature based Automatic Speech Recognizer}

Let $f(x|s)$ be the output probability density of feature vector
$x$ in state $S$. The output probability density is defined by (\ref{eq:missingFeatre}),
page~\pageref{eq:missingFeatre} and becomes: 
\begin{equation}
f(x|S)=\sum_{k=1}^{M}P(k|S)f(x_{r}|k,S),
\end{equation}
 where $M$ is the dimensionality of the Gaussian mixture, and $x_{r}$
are the reliable features in $x$. This means that only reliable features
are used in probability calculation, and thus the recognizer can avoid
undesirable effects due to unreliable features. We used two speech
recognizers. The first one is based on the CASA Tool Kit (CTK)~\cite{barker2001}
hosted at Sheffield University, U.K.\footnote{http://www.dcs.shef.ac.uk/research/groups/spandh/projects/respite/ctk/}
and the second on is the Julius open-source Japanese ASR~\cite{Lee2001}
that we extended to support the above decoding process\footnote{http://julius.sourceforge.jp/}.
According to our preliminary experiments with these two recognizers,
CTK provides slightly better recognition accuracy, while Julius runs
much faster.

\section{Results\label{sec:Results}}

\begin{figure}
\begin{center}\includegraphics[width=0.4\columnwidth,keepaspectratio]{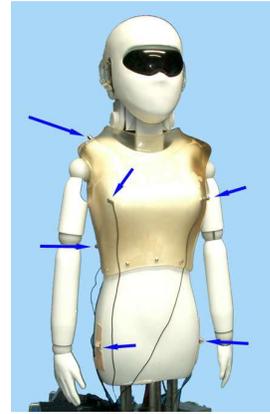}\end{center}

\caption{SIG 2 robot with eight microphones (two are occluded).\label{cap:SIG-2-array}}
\end{figure}

\begin{figure}
\begin{center}\includegraphics[width=5cm,keepaspectratio]{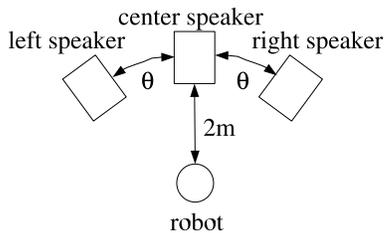}\end{center}

\caption{Position of the speakers relative to the robot in the experimental
setup.\label{cap:Experimental-setup}}
\end{figure}

Our system is evaluated on the SIG2 humanoid robot, on which eight
omni-directional (for the system to work in all directions) microphones
are installed as shown in Fig. \ref{cap:SIG-2-array}. The microphone
positions are constrained by the geometry of the robot because the
system is designed to be fitted on any robot. All microphones are
enclosed within a 22~cm $\times$ 17~cm $\times$ 47~cm bounding
box. To test the system, three Japanese speakers (two males, one female)
are recorded simultaneously: one in front, one on the left, and one
on the right. In nine different experiments, the angle between the
center speaker and the side speakers is varied from 10 degrees to
90 degrees. The speakers are placed two meters away from the robot,
as shown in Fig. \ref{cap:Experimental-setup}. The distance between
the speakers and the robot was not found to have a significant impact
on the performance of the system. The only exception is for short
distances (<50 cm) where performance decreases due to the far-field
assumption we make in this particular work. The position of the speakers
used for the GSS algorithm is computed automatically using the algorithm
described in \cite{ValinICRA2004}.

The room in which the experiment took place is 5~m $\times$ 4~m
and has a reverberation time ($-60\:\mathrm{dB}$) of approximately
$0.3$ seconds. The post-filter parameter $\alpha=1$ (corresponding
to a short-term spectral amplitude (STSA) MMSE estimator) is used
since it was found to maximize speech recognition accuracy\footnote{The difference between $\alpha=1$ and $\alpha=2$ on a subset of
the test set was less than one percent in recognition rate}. When combined together, the GSS, post-filter and missing feature
mask computation require 25\% of a 1.6 GHz Pentium-M to run in real-time
when three sources are present\footnote{Source code for part of the proposed system will be available at http://manyears.sourceforge.net/}.
Speech recognition complexity is not reported as it usually varies
greatly between different engine and settings.

\subsection{Separated Signals}

Spectrograms showing separation of the three speakers\footnote{Audio signals and spectrograms for all three sources are available
at: \texttt{http://www.gel.usherbrooke.ca/laborius/projects/\ Audible/sap/}} are shown in Fig. \ref{cap:Spectrogram-for-separation}, along with
the corresponding mask for static features. Even though the task involves
non-stationary interference with the same frequency content as the
signal of interest, we observe that our post-filter is able to remove
most of the interference. Informal subjective evaluation has confirmed
that the post-filter has a positive impact on both quality and intelligibility
of the speech. This is confirmed by improved recognition results.

\subsection{Speech Recognition Accuracy}

We report speech recognition experiments obtained using the CTK toolkit.
Isolated word recognition on Japanese words is performed using a triphone
acoustic model. We use a speaker-independent 3-state model trained
on 22 speakers (10 males, 12 females), not present in the test set.
The test set includes 200 different ATR phonetically-balanced isolated
Japanese words (300 seconds) for each of the three speakers and is
used on a 200-word vocabulary (each word spoken once). Speech recognition
accuracy on the clean data (no interference, no noise) varies between
94\% and 99\%.

\begin{figure}
\begin{center}\includegraphics[width=0.85\columnwidth,keepaspectratio]{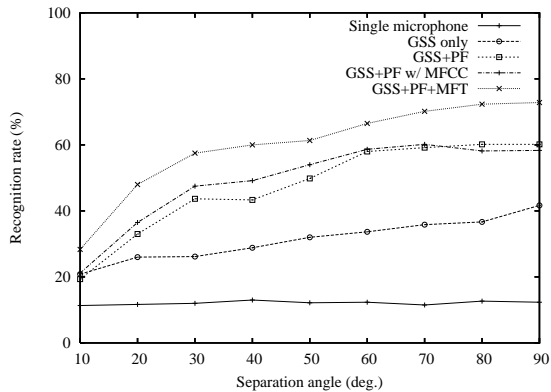}\end{center}

\caption{Speech recognition accuracy results for intervals ranging from ${10}^{\circ}$
to ${90}^{\circ}$ averaged over the three speakers.\label{fig:Speech-recognition-accuracy}}
\end{figure}

Speech recognition accuracy results are presented for five different
conditions:
\begin{enumerate}
\item Single-microphone recording
\item Geometric Source Separation (GSS) only; 
\item GSS with post-filter (GSS+PF); 
\item GSS with post-filter using MFCC features (GSS+PF w/ MFCC)
\item GSS with post-filter and missing feature mask (GSS+PF+MFT). 
\end{enumerate}
Results are shown in Fig. \ref{fig:Speech-recognition-accuracy} as
a function of the angle between sources and averaged over the three
simultaneous speakers. As expected, the separation problem becomes
more difficult as sources are located closer to each other because
the difference in the transfer functions becomes smaller. We find
that the proposed system (GSS+PF+MFT) provides a reduction in relative
error rate compared to GSS alone that ranges from 10\% to 55\%, with
an average of 42\%. The post-filter provides an average of 24\% relative
error rate reduction over use of GSS alone. The relative error rate
reduction is computed as the difference in errors divided by the number
of errors in the reference setup. The results of the post-filter with
MFCC features (4) are included to show that the use of mel spectral
features only has a small effect on the ASR accuracy.

While they seem poor, the results with GSS only can be explained by
the highly non-stationary interference coming from the two other speakers
(especially when the speakers are close to each other) and the fact
that the microphones' placement is constrained by the robot dimensions.
The single microphone results are provided only as a baseline. The
results are very low because a single omni-directional microphone
does not have any acoustic directivity.

\begin{figure}
\begin{center}\includegraphics[width=0.85\columnwidth,keepaspectratio]{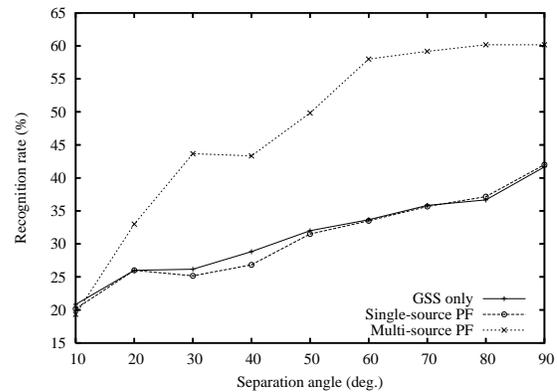}\end{center}

\caption{Effect of the multi-source post-filter on speech recognition accuracy.\label{fig:Effect-of-postfilter}}
\end{figure}

In Fig. \ref{fig:Effect-of-postfilter} we compare the accuracy of
the multi-source post-filter to that of a ``classic'' (single-source)
post-filter that removes background noise but does not take interference
from other sources into account ($\eta=0$). Because the level of
background noise is very low, the single-source post-filter has almost
no effect and most of the accuracy improvement is due to the multi-source
version of the post-filter, which can effectively remove part of the
interference from the other sources. The proposed multi-source post-filter
was also shown in \cite{ValinICASSP2004} to be more effective for
multiple sources than multi-channel approach in \cite{CohenArray2002}.

\section{Conclusion\label{sec:Discussion}}

In this paper we demonstrate a complete multi-microphone speech recognition
system capable of performing speech recognition on three simultaneous
speakers. The system closely integrates all stages of source separation
and missing features recognition so as to maximize accuracy in the
context of simultaneous speakers. We use a linear source separator
based on a simplification of the geometric source separation algorithm.
The non-linear post-filter that follows the initial separation step
is a short-term spectral amplitude MMSE estimator. It uses a background
noise estimate as well as information from all other sources obtained
from the geometric source separation algorithm.

In addition to removing part of the background noise and interference
from other sources, the post-filter is used to compute a missing feature
mask representing the reliability of mel spectral features. The mask
is designed so that only spectral regions dominated by interference
are marked as unreliable. When compared to the GSS alone, the post-filter
contributes to a 24\% (relative) reduction in the word error rate
while the use of the missing feature theory-based modules yields a
reduction of 42\% (also when compared to GSS alone). The approach
is specifically designed for recognition on multiple sources and we
did not attempt to improve speech recognition of a single source with
background noise. In fact, for a single sound source, the proposed
work is strictly equivalent to commonly used single-source techniques.

We have shown that robust simultaneous speakers speech recognition
is possible when combining the missing feature framework with speech
enhancement and source separation with an array of eight microphones.
To our knowledge, there is no work reporting multi-speaker speech
recognition using missing feature theory. This is why this paper is
meant more as a proof of concept for a complete auditory system than
a comparison between algorithms for performing specific signal processing
tasks. Indeed, the main challenge here is the adaptation and integration
of the algorithms on a mobile robot so that the system can work in
a real environment (moderate reverberation) and that real-time speech
recognition with simultaneous speakers be possible.


In future work, we plan to perform the speech recognition with moving
speakers and adapt the post-filter to work even in highly reverberant
environments, in the hope of developing new capabilities for natural
communication between robots and humans. Also, we have shown that
the cepstral-domain speech recognition usually performs slightly better,
so it would be desirable for the technique to be generalized to the
use of cepstral features instead of spectral features.

\section*{Acknowledgement}
Jean-Marc Valin was supported by the Natural Sciences and Engineering Research Council of Canada (NSERC), the Quebec Fonds de recherche sur la nature et les technologies and the JSPS short-term exchange student scholarship. Jean Rouat is supported by NSERC. Fran\c{c}ois Michaud holds the Canada Research Chair (CRC) in Mobile Robotics and Autonomous Intelligent Systems. This research is supported financially by the CRC Program, NSERC and the Canadian Foundation for Innovation (CFI). This research was also partially supported by MEXT and JSPS, Grant-in-Aid for Scientific Research (A) No.15200015 and Informatics No.16016251, and Informatics Research Center for Development of Knowledge Society Infrastructure (COE program of MEXT, Japan).

\bibliographystyle{IEEEtran}
\bibliography{iros,BiblioAudible,pub2,localize}

\begin{biography}[{\includegraphics[width=1in,keepaspectratio]{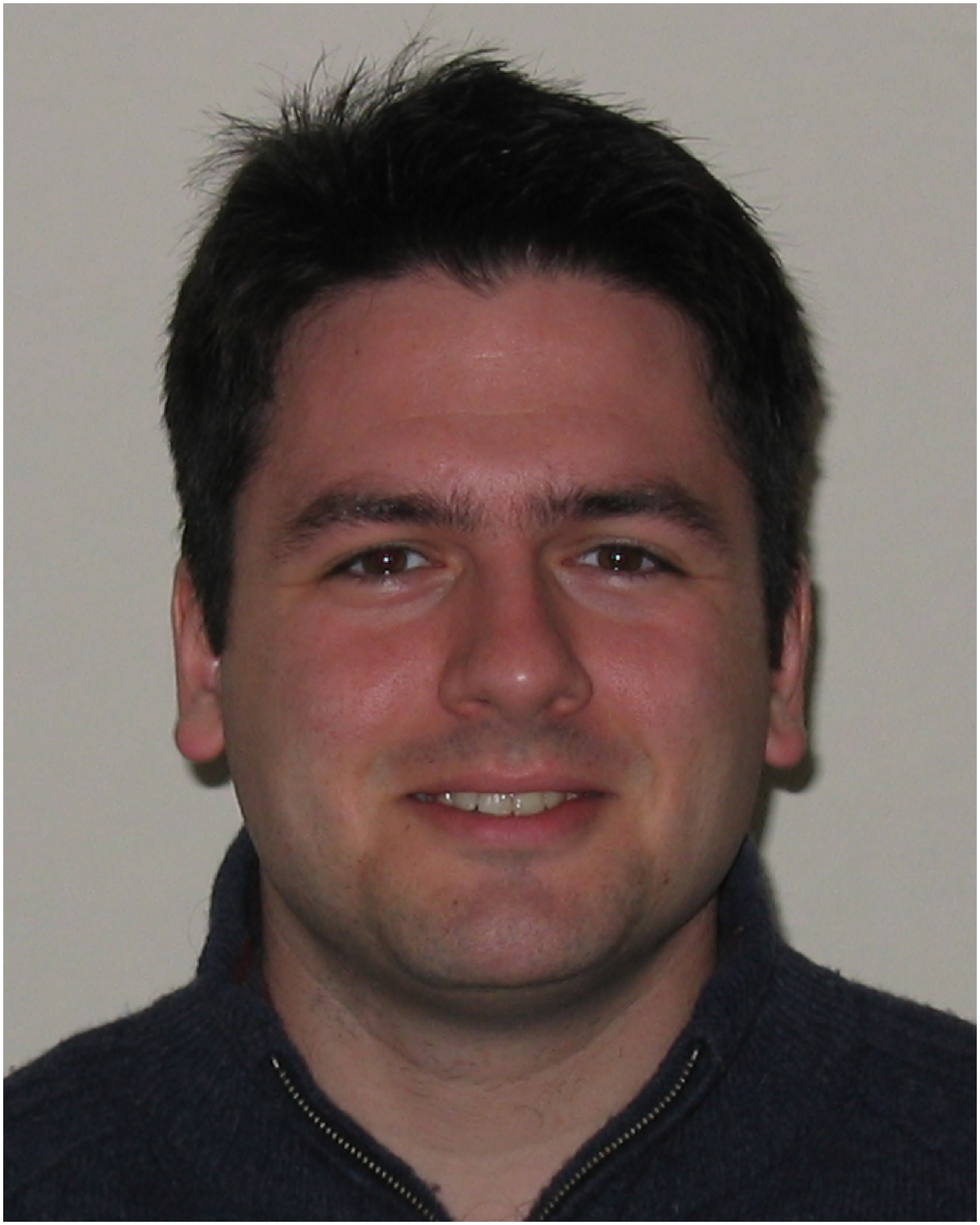}}]{Jean-Marc
Valin} (S'03-M'05) holds B.Eng. ('99), M.A.Sc. ('01) and Ph.D. ('05)
degrees in electrical engineering from the University of Sherbrooke.
His Ph.D. research focused on bringing auditory capabilities to a
mobile robotics platform, including sound source localization and
separation. Prior to his PhD, Jean-Marc has also worked in the field
of automatic speech recognition. In 2002, he wrote the Speex open
source speech codec, which he keeps maintaining to this date. Since
2005, he is a postdoctoral fellow at the CSIRO ICT Centre in Sydney,
Australia. His research topics include acoustic echo cancellation
and microphone array processing. He is a member of the IEEE signal
processing society.\end{biography}

\begin{biography}[{\includegraphics[width=1in,keepaspectratio]{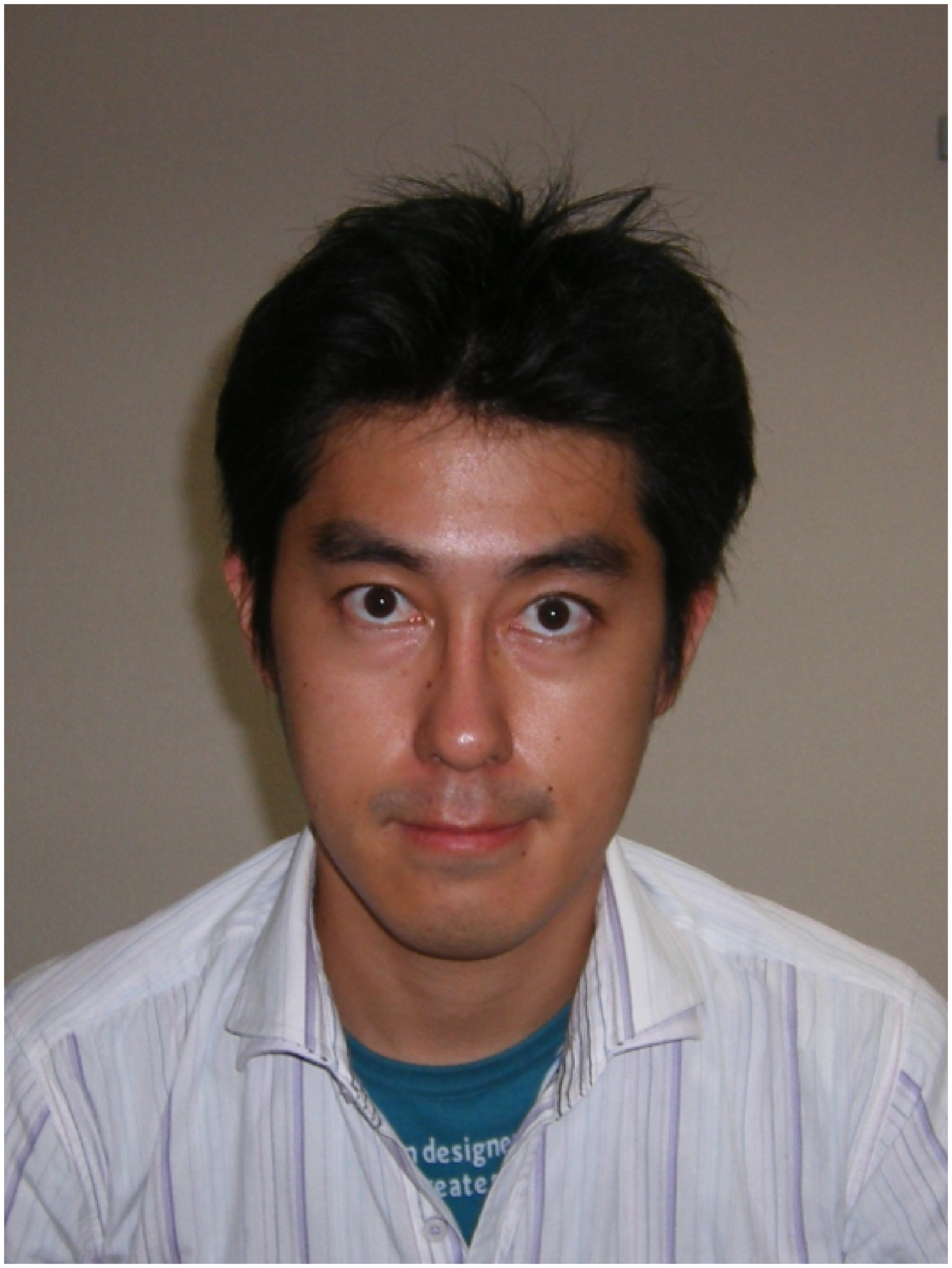}}]{Shun'ichi
Yamamoto} received the B.S. and M.S. degrees from Kyoto University,
Kyoto, Japan, in 2003 and 2005, respectively. He is currently pursuing
the Ph.D degree in the Department of Intelligence Science and Technology,
Graduate School of Informatics, Kyoto University. His research interests
include automatic speech recognition, sound source separation, sound
source localization for robot audition. Mr. Yamamoto is a member of
IEEE. He has received several awards including the IEEE Robotics and
Automation Society Japan Chapter Young Award.\end{biography} 

\begin{biography}[{\includegraphics[width=1in,keepaspectratio]{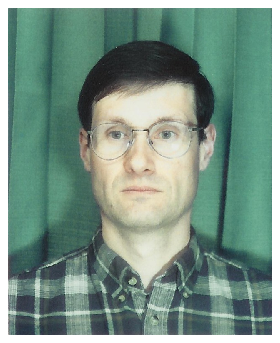}}]{Jean
Rouat} holds a master degree in Physics from Univ. de Bretagne, France
(1981), an E. \& E. master degree in speech coding and speech recognition
from Université de Sherbrooke (1984) and an E. \& E. Ph.D. in cognitive
and statistical speech recognition jointly with Université de Sherbrooke
and McGill University (1988). From 1988 to 2001 he was with Université
du Québec à Chicoutimi (UQAC). In 1995 and 1996, he was on a sabbatical
leave with the Medical Research Council, Applied Psychological Unit,
Cambridge, UK and the Institute of Physiology, Lausanne, Switzerland.
In 1990 he founded the ERMETIS, Microelectronics and Signal Processing
Research Group from UQAC. From September 2006 to March 2007 he was
with McMaster University with the ECE department. He is now with Université
de Sherbrooke where he founded the Computational Neuroscience and
Intelligent Signal Processing Research group. Since February 2007
he is also invited professor in the biological sciences dept from
Université de Montréal. His research interests cover audition, speech
and signal processing in relation with networks of spiking neurons.
He regularly acts as a reviewer for speech, neural networks and signal
processing journals. He is an active member of scienti?c associations
(Acoustical Society of America, Int. Speech Communication, IEEE, Int.
Neural Networks Society, Association for Research in Otolaryngology,
etc.). He is a senior member of the IEEE and was on the IEEE technical
committee on Machine Learning for Signal Processing from 2001 to 2005.\end{biography}

\begin{biography}[{\includegraphics[width=1in,keepaspectratio]{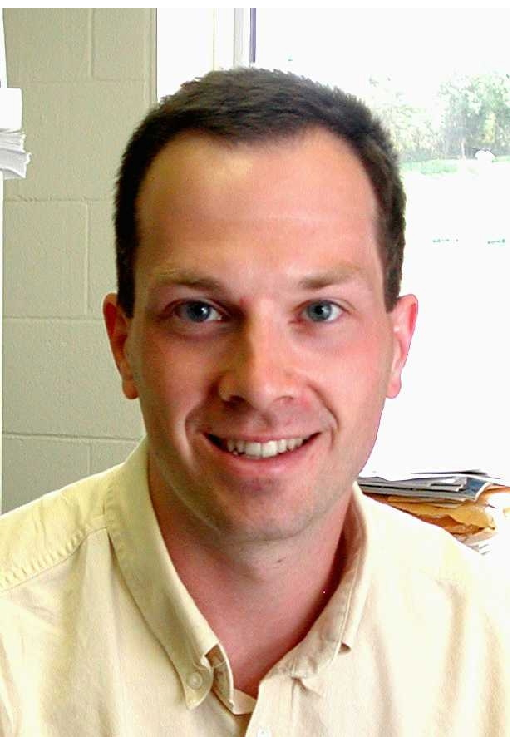}}]{François
Michaud} (M'90) received his bachelor's degree ('92), Master's degree
('93) and Ph.D. degree ('96) in electrical engineering from the Université
de Sherbrooke, Québec Canada. After completing postdoctoral work at
Brandeis University, Waltham MA ('97), he became a faculty member
in the Department of Electrical Engineering and Computer Engineering
of the Université de Sherbrooke, and founded LABORIUS, a research
laboratory working on designing intelligent autonomous systems that
can assist humans in living environments. His research interests are
in architectural methodologies for intelligent decision-making, design
of autonomous mobile robotics, social robotics, robot for children
with autism, robot learning and intelligent systems. Prof. Michaud
is the Canada Research Chairholder in Autonomous Mobile Robots and
Intelligent Systems. He is a member of IEEE, AAAI and OIQ (Ordre des
ingénieurs du Québec). In 2003 he received the Young Engineer Achievement
Award from the Canadian Council of Professional Engineers.\end{biography} 

\begin{biography}[{\includegraphics[width=1in,keepaspectratio]{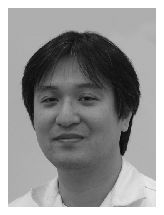}}]{Kazuhiro
Nakadai} received B.E. in electrical engineering in 1993, M.E. in
information engineering in 1995 and Ph.D. in electrical engineering
in 2003, from the University of Tokyo. He worked for Nippon Telegraph
and Telephone and NTT Comware Corporation from 1995 to 1999. He was
a researcher for Kitano Symbiotic Systems Project, ERATO, Japan Science
and Technology Corporation from 1999 to 2003. He is currently a senior
researcher in Honda Research Institute Japan, Co., Ltd. Since 2006,
he is also Visiting Associate Professor at Tokyo Institute of Technology.
His research interests include signal and speech processing, AI and
robotics, and he is currently engaged in computational auditory scene
analysis, multi-modal integration and robot audition. He received
the best paper award at IEA/AIE-2001 from the International Society
for Applied Intelligence, and the best paper finalist of International
Conference on Intellignet Robots and Systems (IROS 2001). He is a
member of RSJ, JSAI, ASJ, and IEEE.\end{biography}

\begin{biography}[{\includegraphics[width=1in,keepaspectratio]{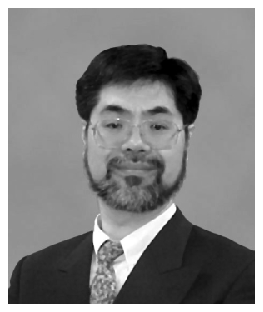}}]{Hiroshi
G. Okuno} Hiroshi G. Okuno received the B.A. and Ph.D degrees from
the University of Tokyo, Tokyo, Japan, in 1972 and 1996, respectively.
He worked for Nippon Telegraph and Telephone, Kitano Symbiotic Systems
Project, and Tokyo University of Science. He is currently a Professor
in the Department of Intelligence Science and Technology, Graduate
School of Informatics, Kyoto University, Kyoto, Japan. He was a Visiting
Scholar at Stanford University, Stanford, CA, and Visiting Associate
Professor at the University of Tokyo. He has done research in programming
languages, parallel processing, and reasoning mechanisms in AI, and
is currently engaged in computational auditory scene analysis, music
scene analysis, and robot audition. He edited (with D. Rosenthal)
Computational Auditory Scene Analysis (Princeton, NJ: Lawrence Erlbaum,
1998) and (with T. Yuasa) Advanced Lisp Technology (London, U.K.:
Taylor \&Francis, 2002). Dr. Okuno has received various awards including
the 1990 Best Paper Award of JSAI, the Best Paper Award of IEA/AIE-2001
and 2005, and IEEE/RSJ Nakamura Award for IROS-2001 Best Paper Nomination
Finalist. He was also awarded 2003 Funai Information Science Achievement
Award. He is a member of the IPSJ, JSAI, JSSST, JSCS, RSJ, ACM, AAAI,
ASA, and ISCA.\end{biography}
\end{document}